\documentclass[oneside,a4paper,draft]{article}

\newif\ifdraft\draftfalse

\newif\ifreva\revafalse
\usepackage{graphicx}
\usepackage{epstopdf}%
\usepackage{subfigure}%
\usepackage{amsthm,amsmath,amssymb}%
\usepackage{url}%
\usepackage{fullpage}

\usepackage[style=alphabetic,natbib=true,isbn=false]{biblatex}
\theoremstyle{plain}%
\newtheorem{theorem}{Theorem}[section]

\theoremstyle{definition}
\newtheorem{definition}[theorem]{Definition}

\theoremstyle{remark}

\newtheorem{example}{Example}

\usepackage{booktabs}

\ifreva
\usepackage{color}
\newcommand{\revamark}[0]{\color{blue}}
\newcommand{\reva}[1]{{\revamark #1}}
\else
\newcommand{\revamark}[0]{}
\newcommand{\reva}[1]{{#1}}
\fi

\ifdraft
\usepackage{showlabels}
\usepackage{todonotes}
\else
\usepackage[disable]{todonotes}
\fi

\usepackage{amsmath}
\usepackage{paralist}
\usepackage[T1]{fontenc}
\usepackage[utf8]{inputenc}
\usepackage{multirow}

\usepackage{fancyvrb}
\fvset{numbers=left,numberblanklines=false,frame=single,commentchar=?}

\makeatletter
\def\label@in@display{\gdef\df@label}
\makeatother

\makeatletter
\newcommand{\crasp}{\textsf{CaspR}}
\newcommand{\ilv}[1]{\ensuremath{\mi{#1}}}
\newcommand{\objv}{{\tt[V]}}
\newcommand{\obju}{{\tt[U]}}
\newcommand{\objva}{{\tt[VA]}}
\newcommand{\objua}{{\tt[UA]}}

\newcommand{\clasp}{\textsf{Clasp}}
\newcommand{\wasp}{\textsf{Wasp}}
\newcommand{\clingo}{\textsf{Clingo}}
\newcommand{\gringo}{\textsf{Gringo}}

\newcommand{\runlim}{\textsf{runlim}}
\newcommand{\leanparagraph}[1]{\smallskip\noindent\textbf{#1.}}

\newcommand{\quo}[1]{\text{`#1'}}
\newcommand{\bit}{\begin{itemize}}
\newcommand{\eit}{\end{itemize}}
\newcommand{\bci}{\begin{compactitem}}
\newcommand{\eci}{\end{compactitem}}
\newcommand{\bipe}[1]{\begin{inparaenum}[#1]}
\newcommand{\eipe}{\end{inparaenum}}
\newcommand{\ba}{\begin{array}}
\newcommand{\ea}{\end{array}}
\newcommand{\beq}{\begin{equation}}
\newcommand{\eeq}[1]{\label{#1}\end{equation}}
\newcommand{\mi}[1]{\ensuremath{\mathit{#1}}}
\newcommand{\mc}[1]{\ensuremath{\mathcal{#1}}}
\newcommand{\mb}[1]{\ensuremath{\mathbf{#1}}}

\newcommand{\lors}{\,{\lor}\,}
\newcommand{\ins}{\,{\in}\,}

\newcommand{\gts}{\,{>}\,}
\newcommand{\ges}{\,{\ge}\,}
\newcommand{\les}{\,{\le}\,}
\newcommand{\lts}{\,{<}\,}
\newcommand{\eqs}{\,{=}\,}
\newcommand{\pluss}{\,{+}\,}
\newcommand{\neqs}{\,{\neq}\,}

\newcommand{\cups}{\,{\cup}\,}
\newcommand{\caps}{\,{\cap}\,}

\newcommand{\subseteqs}{\,{\subseteq}\,}

\def\cC{\mc{C}}

\def\cT{\mc{T}}
\def\cX{\mc{X}}

\def\HBP{\mi{H\!B}_P}

\newcommand{\bbN}{\ensuremath{\mathbb{N}}}
\newcommand{\cols}{\,{:}\,}
\newcommand{\grnd}{\ensuremath{\mi{grnd}}}
\def\fP{fP}

\newcommand{\lar}{\ensuremath{\leftarrow}}
\newcommand{\lars}{\,{\lar}\,}
\newcommand{\naf}{\mb{not}\,}
\newcommand{\wlar}{\ensuremath{\reflectbox{\ensuremath{\leadsto}}}}
\newcommand{\wlars}{\,{\wlar}\,}
\newcommand{\arspc}{\hphantom{\lars}}

\newcommand{\mytableObjectiveConstraints}[0]{
\begin{table}
\caption{Objective functions and (weak) constraints that are used to realize them in MM and CM encodings.}
\smallskip
\centering
\begin{tabular}{@{\quad}l@{\qquad\quad}cccc@{\qquad\quad}cccc@{\quad}}
\toprule
& \multicolumn{4}{@{}c@{\qquad\qquad}}{Encoding MM}
& \multicolumn{4}{@{}c@{\qquad}}{Encoding CM} \\
(Weak) constraint
  & \objv & \obju & \objva & \objua & \objv & \obju & \objva & \objua \\
\midrule
\eqref{eqMMOmitCost} = \eqref{eqCMOmitCost}
  &   X   &   X   &   X    &   X    &   X   &   X   &   X    &   X    \\
\eqref{eqMMUseCost} = \eqref{eqCMUseCost}
  &   X   &       &   X    &        &   X   &       &   X    &        \\
\eqref{eqMMTransForbid}
  &   X   &   X   &        &        &       &       &        &        \\
\eqref{eqMMTransCost}
  &       &       &   X    &   X    &       &       &        &        \\
\eqref{eqCMTransForbid}
  &       &       &        &        &   X   &   X   &        &        \\
\eqref{eqCMTransCost}
  &       &       &        &        &       &       &   X    &   X    \\
\bottomrule%
\end{tabular}%
\label{tblObjectiveConstraints}%
\end{table}%
}

\newcommand{\myexptable}[0]{
\def\mymcc{\multicolumn{3}{@{~}l@{}}}
\begin{table}
\caption{Experimental results, accumulated over ASP tools, encodings, and objective functions, and accumulated over eight practically relevant use cases.}
\smallskip
\centering
\begin{tabular}{@{~}c@{\,}c@{\,}c@{}r@{~~}r@{~~}r@{~~}r@{~}r@{~~}r@{~~}r@{~~}r@{~~}r@{~~}r@{~~}r@{~~}r@{~}}
\toprule
&&& MO & TO & SAT & OPT & OPT & $T$ & $M$ & $T_\mi{grd}$ & $\mi{Opt}$ & $\mi{Opt}$ & $\mi{Chc}$ & $\mi{Cnf}$ \\
\mymcc{Accumulation}
 &  \# &  \# &  \# &  \# & \% & sec & MB & sec & avg & max & \# & \# \\
\midrule
\mymcc{\clingo}
 &        20 &     0 & {\bf 111} & {\bf 301} & {\bf 70} &      96 &      738 &      5.7 \\
\mymcc{\gringo+\clasp}
 &        20 &     0 & {\bf 111} & {\bf 301} & {\bf 70} &      96 & {\bf 695}&      2.7 \\
\mymcc{\gringo+\wasp}
 &        34 &    81 &       89  &       228 &      53  &     145 &     1075 &      2.6 \\
\midrule
\mymcc{CM}
 & {\bf 4}&     0 &     87 &      125  &      58  &    137 &      567 &      3.7 &      6.8 &    368.9 &   1M &  331K \\
\mymcc{MM}
 &     16 &     0 &     24 & {\bf 176} & {\bf 82} &     54 &      822 & {\bf 1.6}&{\bf 1.0} &{\bf 84.7}& 132K &    5K \\
\midrule
\mymcc{\obju}
 &      2 &     0 &     38 &     68 &     63  &     125 & {\bf 398}& {\bf 1.8}&      1.2 &     15.2 &   2M &  538K \\
\mymcc{\objv}
 &      2 &     0 &      3 &    103 &     95  & {\bf 24}& {\bf 397}& {\bf 1.8}&      0.0 &      0.0 & 177K &   22K \\
\mymcc{\objua}
 &     8 &     0 &     58 &     42 &     39  &     177 &      994 &      3.5 &     14.4 &    368.9 & 606K &   99K \\
\mymcc{\objva}
 &     8 &     0 &     12 &     88 &     81  & {\bf 58}&      990 &      3.5 &      0.0 &      0.6 & 132K &   13K \\
\midrule
{\bf{}DS1} & {\bf{}\obju} & {\bf{}CM}
 & 0 &     0 &{\bf 21}&      0 &       0 &     300 &      384 &      2.8 & {\bf 5.8}&{\bf 15.2}&   5M &    2M \\
{\bf{}DS1} & {\bf{}\obju} & {\bf{}MM}
 & 2 &     0 &      1 &     18 &      86 &      36 &      806 &      1.0 &        0 &        0 &   8K &   314 \\
{\bf{}DS1} & {\bf{}\objua} & {\bf{}CM}
 & 2 &     0 &     19 &      0 &       0 &     282 &     1474 &      8.3 &     37.1 &    368.9 &   1M &  338K \\
{\bf{}DS1} & {\bf{}\objua} & {\bf{}MM}
 & 4 &     0 &      8 &      9 &      43 &     136 &     1206 &      1.0 &      5.9 &     84.7 & 458K &   38K \\
DS2 & \objv & CM
 & 0 &     0 &      2 &     31 &      94 &      35 &      166 &      1.6 &      0.0 &      0.0 & 465K &   69K \\
{\bf{}DS2} & {\bf{}\objv} & {\bf{}MM}
 & 0 &     0 &      0 &{\bf 33}&{\bf 100}& {\bf 13}&      376 &      1.9 &        0 &        0 &  70K &    35 \\
DS2 & \objva & CM
 & 0 &     0 &      7 &     26 &      79 &      87 &      506 &      3.7 &      0.0 &      0.6 & 323K &   39K \\
DS2 & \objva & MM
 & 2 &     0 &      5 &     26 &      79 &      64 &     1034 &      2.0 &      0.0 &      0.0 &  84K &   134 \\
\bottomrule
\end{tabular}%
\label{tblExp}%
\end{table}%
}

\newcommand{\myencodingfigurecostcommonMath}[0]{%
\begin{figure}
\begin{align}%
\mi{na(N)} &\lars \mi{N \eqs \#count\,\{\,A\,{:}\,mention(A,\_,\_,\_)\,\}.} \label{eqNofAnno} \\
\mi{acmen(A,Am,mid(S,E))} &\lars \mi{mention(A,Am,S,E).} \label{eqAcmen} \\
\mi{asamechain(A,Am_1,Am_2)} &\lars \mi{cm(A,C,Am_1),\, cm(A,C,Am_2),\, Am_1 \neqs Am_2.} \label{eqASameChain} \\
\mi{csamechain(M_1,M_2)}
  &\lars \mi{acmen(A,Am_1,M_1),\,acmen(A,Am_2,M_2),} \notag \\
  &\arspc \mi{asamechain(A,Am_1,Am_2),\,M_1 \lts M_2.}
   \label{eqCSameChain} \\
\mi{evidence(M_1,M_2,Ev)}
  &\lars \mi{csamechain(M_1,M_2),\,Ev \eqs \#count\,\{} \notag \\
  &\arspc \ \mi{A\,{:}\,asamechain(A,Am_1,Am_2),\,Am_1 \lts Am_2,} \notag \\
  &\arspc \ \mi{\hphantom{F\,{:}\,}acmen(A,Am_1,M_1),\,acmen(A,Am_2,M_2)} \notag \\
  &\arspc \mi{\},\, Ev \gts 0}. \label{eqPosEvidence} \\
\mi{cmomitcost(M_1,M_2,2\,{\cdot}\,K)}
  &\lars \mi{csamechain(M_1,M_2),\,evidence(M_1,M_2,K).} \label{eqOmitCost} \\
\mi{cmusecost(M_1,M_2,N\,{-}\,K)}
  &\lars \mi{csamechain(M_1,M_2),\,evidence(M_1,M_2,K),} \notag \\
  &\arspc \mi{N\,{-}\,K \gts 0,\, na(N).} \label{eqUseCost}
\end{align}%
\caption{Common deterministic definitions used in objective function constraints.}%
\label{figCommon}%
\end{figure}%
}

\newcommand{\myencodingfigureenforceMath}[0]{%
\begin{figure}
\begin{gather}%
\begin{align}%
  \mi{force_\mi{tok}(Token)} &\lars \mi{empty(Token).} \label{eqForceEmpty} \\
  \mi{force_\mi{tok}(S)} &\lars \mi{mention(forced,\_,S,E).} \label{eqForceStart} \\
  \mi{force_\mi{tok}(E)} &\lars \mi{mention(forced,\_,S,E).} \label{eqForceEnd} \\
  \mi{force_\mi{same}(M_1,M_2)}
    &\lars \mi{cm(forced,C,Am_1),\, cm(forced,C,Am_2),\, Am_1 \neqs Am_2,\, M_1 \lts M_2,} \notag \\
    &\arspc \mi{acmen(forced,Am_1,M_1),\, acmen(forced,Am_2,M_2).} \label{eqForceSame} \\
  \mi{force_\mi{diff}(M_1,M_2)}
    &\lars \mi{cm(forced,C_1,Am_1),\, cm(forced,C_2,Am_2),\, C_1 \lts C_2,} \notag \\
    &\arspc \mi{acmen(forced,Am_1,M_1),\, acmen(forced,Am_2,M_2).} \label{eqForceDiff}
\end{align} \\
\begin{align}
  &\lars \mi{force_\mi{tok}(S),\, result_{cm}(\_,mid(S,E)),
    \, \naf\, mention(forced,\_,S,E).} \label{eqForceNoInventS} \\
  &\lars \mi{force_\mi{tok}(E),\, result_{cm}(\_,mid(S,E)),
    \, \naf\, mention(forced,\_,S,E).} \label{eqForceNoInventE} \\
  &\lars \mi{mention(forced,\_,S,E),\, \naf\ result_{cm}(\_,mid(S,E)).} \label{eqForceReproduce} \\
  &\lars \mi{force_\mi{same}(M_1,M_2),\, result_{cm}(C_1,M_1),\, result_{cm}(C_2,M_2),\, C_1 \neqs C_2.} \label{eqForceSameCheck} \\
  &\lars \mi{force_\mi{diff}(M_1,M_2),\, result_{cm}(C,M_1),\, result_{cm}(C,M_2).} \label{eqForceDiffCheck}
\end{align}%
\end{gather}%
\caption{%
  Program module for semi-automatic mode that allows for enforcing certain mentions and chains, moreover empty tokens, i.e., tokens without coreference information.}%
\label{figEnforceMath}%
\end{figure}%
}

\newcommand{\myencodingfigurecommonLinkMath}[0]{%
\begin{figure}%
\revamark
\begin{align}
\mi{link(A,Am_1,Am_2)} &\lars \mi{cm(A,C,Am_1),\,cm(A,C,Am_2),\,Am_1 \lts Am_2.} \label{eqLinkFromACM} \\
\mi{\{\,uselink(A,Am_1,Am_2)\, \}} &\lars \mi{link(A,Am_1,Am_2).} \label{eqGuessUseLink} \\
\mi{clink(M_1,M_2)} &\lars \mi{uselink(A,Am_1,Am_2),\,M_1 \lts M_2,} \notag \\
&\arspc \mi{acmen(A,Am_1,M_1),\, mention(A,Am_,M_2).} \label{eqCanonicalize1} \\
\mi{clink(M_1,M_2)} &\lars \mi{uselink(A,Am_2,Am_1),\,M_1 \lts M_2,} \notag \\
&\arspc \mi{acmen(A,Am_1,M_1),\, mention(A,Am_2,M_2).} \label{eqCanonicalize2}
\end{align}%
\caption{Common link selection and canonicalization encoding.}%
\label{figCommonLink}%
\end{figure}
}

\newcommand{\myencodingfiguremmMath}[0]{%
\begin{figure}%
\begin{align}
\mi{cc(M_1,M_2)} &\lars \mi{clink(M_1,M_2).} \label{eqMMCCCopy} \\
\mi{cc(M_2,M_1)} &\lars \mi{cc(M_1,M_2).} \label{eqMMCCReflexive} \\
\mi{cc(M_1,M_3)} &\lars \mi{cc(M_1,M_2),\,cc(M_2,M_3).} \label{eqMMCCTransitive} \\
\mi{notrepresentative(M_2)} &\lars \mi{cc(M_1,M_2),\, M_1 \lts M_2.} \label{eqMMNotRep} \\
\mi{result_{cm}(M_1,M_2)} &\lars \mi{cc(M_1,M_2),\, \naf notrepresentative(M_1).} \label{eqMMResultCM} \\
\mi{resultchain(C)} &\lars \mi{result_{cm}(C,\_).} \label{eqMMResultC}
\end{align}%
\caption{MM encoding.}%
\label{figMM}%
\end{figure}
\begin{figure}%
\begin{align}
&\wlars \naf \mi{clink(M_1,M_2), cmomitcost(M_1,M_2,Cost).}
  & \mi{[Cost@1,M_1,M_2,omit]}& \label{eqMMOmitCost} \\
&\wlars \mi{clink(M_1,M_2), cmusecost(M_1,M_2,Cost).}
  & \mi{[Cost@1,M_1,M_2,use]}& \label{eqMMUseCost} \\
&\lars \mi{cc(M_1,M_2), M_1 \lts M_2, \naf clink(M_1,M_2).} \label{eqMMTransForbid} \\
&\wlars \mi{cc(M_1,M_2), M_1 \lts M_2, \naf clink(M_1,M_2).}
  & \mi{[1@1,M_1,M_2]}& \label{eqMMTransCost}
\end{align}%
\caption{MM objective function variations.}%
\label{figMMObjective}%
\end{figure}%
}

\newcommand{\myencodingfigurecmMath}[0]{%
\begin{figure}%
\begin{gather}
\begin{align}
\mi{countchain(A,N)} &\lars \mi{cm(A,\_,\_),}\notag \\
  &\arspc \mi{N \eqs \#count\,\{\,C\,{:}\,cm(A,C,\_)\,\}.}
  \label{eqCMCountChain} \\
\mi{maxchain(N)} &\lars \mi{N \eqs \#max\, \{\, C\, {:}\, countchain(\_,C)\, \}.} \label{eqCMMaxChain} \\
\mi{chainlimit(6/5\,{\cdot}\,N)} &\lars \mi{maxchain(N).} \label{eqCMChainLimit} \\
\mi{\{\,resultchain(C)\ {:}\ C \ins \{1,\ldots,N\}\,\}\,}
  &\lars \mi{chainlimit(N).} \label{eqCMResultChain} \\
\mi{resultchain(C_2)} &\lars \mi{resultchain(C_1),\, C_2 \ins \{ 1,\ldots, C_2\,{-}\,1\}.} \label{eqCMResultChainSB} \\ %
\mi{cmention(M)} &\lars \mi{clink(M,\_).} \label{eqCMCm1} \\
\mi{cmention(M)} &\lars \mi{clink(\_,M).} \label{eqCMCm2} \\
\mi{1\,\{\,result_{cm}(C,M)\ {:}\ resultchain(C)\,\}}\,1
  &\lars \mi{cmention(M).} \label{eqCMResultChainMention}
\end{align} \\
\begin{align}
&\lars \mi{clink(M_1,M_2),\, result_{cm}(C,M_1),\, \naf result_{cm}(C,M_2).} & \label{eqCMSync1} \\
&\lars \mi{clink(M_1,M_2),\, \naf result_{cm}(C,M_1),\, result_{cm}(C,M_2).} & \label{eqCMSync2}
\end{align}%
\end{gather}%
\caption{CM encoding.}%
\label{figCM}%
\end{figure}
}

\newcommand{\myencodingfigurecmobjMath}[0]{%
\begin{figure}%
\begin{align}
&\ifmeasuring@\else\omit\fi%
 \wlars \naf \mi{clink(M_1,M_2), cmomitcost(M_1,M_2,Cost).} \notag \\
&\ifmeasuring@\else\omit\fi%
 \hfill\mi{[Cost@1,M_1,M_2,omit]} \label{eqCMOmitCost} \\
&\ifmeasuring@\else\omit\fi%
 \wlars \mi{clink(M_1,M_2), cmusecost(M_1,M_2,Cost).} \notag \\
&\ifmeasuring@\else\omit\fi%
 \hfill\mi{[Cost@1,M_1,M_2,use]} \label{eqCMUseCost} \\
&\lars \mi{result_{cm}(C,M_1),\,result_{cm}(C,M_2),\, M_1 \lts M_2,\,\naf clink(M_1,M_2).} \label{eqCMTransForbid} \\
&\wlars \mi{result_{cm}(C,M_1),\,result_{cm}(C,M_2),\, M_1 \lts M_2,\, \naf clink(M_1,M_2).} \notag \\
&\ifmeasuring@\else\omit\fi%
 \hfill\mi{[1@1,M_1,M_2]} \label{eqCMTransCost}
\end{align}%
\caption{CM objective function variations.}%
\label{figCMObjective}%
\end{figure}%
}

\newcommand{\myencodingfigureCommonConstraints}[0]{%
\begin{figure}%
\revamark
\begin{align}
&\lars \mi{resultchain(C),\,\#count\,\{\,M\,{:}\,result_{cm}(C,M)\,\} \les 1.} \label{eqNoSingletons} \\
&\lars \mi{result_{cm}(C,mid(S_1,E_1)),\, result_{cm}(C,mid(S_2,E_2)),} \notag \\
&\arspc \mi{S_1 \les S_2,\, E_2 \les E_1,\, (S_1,E_1) \neqs (S_2,E_2).} \label{eqNoNestedCoref}
\end{align}%
\caption{Common structural constraints on adjudication solutions.}%
\label{figCommonConstraints}%
\end{figure}%
}

\newcommand{\mydatasettable}{
\def\hph{}
\begin{table}
\centering
\caption{Properties of our real-world datasets which are based on the METU-Sabanci Turkish Treebank.}
\smallskip
\begin{tabular}{@{\quad}l@{\qquad}r@{\quad}r@{\quad}r@{\quad\qquad}r@{\quad}r@{\quad}r@{\quad}}
\toprule
Dataset & \multicolumn{3}{c@{\qquad}}{DS1 ({\bf 21} instances)} & \multicolumn{3}{c}{DS2 ({\bf 33} instances)} \\
\midrule
           &   min &   avg &    max & min   &    avg & max    \\
\midrule
\# Annotators        &   6\hph &   6.5 &    8\hph &   9\hph &   10.3 &   11\hph \\
\# Chains            &  78\hph & 132.7 &  197\hph &  34\hph &  294.6 &  599\hph \\
\# Annotated mentions& 247\hph & {\bf 596.6} & 1289\hph & 117\hph & {\bf 1561.4} & 3897\hph \\
\# Distinct mentions & 169\hph & {\bf 316.7} &  702\hph &  17\hph &  {\bf 159.8} &  358\hph \\
Longest chain        &   7\hph &  33.1 &   66\hph &   6\hph &   28.4 &   70\hph \\
\bottomrule
\end{tabular}
\label{tblDatasets}
\end{table}%
}

\newcommand{\exampleInputsFigure}[0]{%
\def\customspacer{\hspace*{3.5em}}
\begin{figure}%
  \centering%
  \subfigure[in1.conll]{\label{figIn1}%
  ~~\framebox{%
  \BVerbatimInput[numbers=none,commentchar=\#]%
  {examples_in1.conll}}~~%
  }
  \customspacer
  \subfigure[in2.conll]{\label{figIn2}~~\framebox{%
  \BVerbatimInput[numbers=none,commentchar=\#]%
  {examples_in2.conll}}~~}
  \customspacer
  \subfigure[in3.conll]{\label{figIn3}~~\framebox{%
  \BVerbatimInput[numbers=none,commentchar=\#]%
  {examples_in3.conll}}~~}
  \customspacer
  \subfigure[in4.conll]{\label{figIn4}~~\framebox{%
  \BVerbatimInput[numbers=none,commentchar=\#]%
  {examples_in4.conll}}~~}
  \caption{Minimalistic example CoNLL input files containing two columns: token index (first column) and coreference chains (second column).}%
  \label{figInFiles}%
\end{figure}%
}

\newcommand{\exampleOutputsFigureLastColumn}{%
\begin{figure}
  \centering%
  \subfigure[\obju]{\label{figOutu}~~\framebox{%
  \BVerbatimInput[numbers=none]{examples_out.u.conll.cut}}~~}
  \qquad\ \
  \subfigure[\objua]{\label{figOutua}~~\framebox{%
  \BVerbatimInput[numbers=none]{examples_out.ua.conll.cut}}~~}
  \quad\ \ \
  \subfigure[\objv]{\label{figOutv}%
  ~~\framebox{%
  \BVerbatimInput[numbers=none,lastline=6]{examples_out.v.conll.cut}%
  }~~\framebox{%
  \BVerbatimInput[numbers=none,firstline=7]{examples_out.v.conll.cut}%
  }~~%
  }
  \quad\ \
  \subfigure[\objva]{\label{figOutva}~~\framebox{%
  \BVerbatimInput[numbers=none]{examples_out.va.conll.cut}}~~}
  \ \
  \caption{Optimal solutions of automatic adjudication of inputs in Figure~\ref{figInFiles} with all four objectives.}
  \label{figOutFiles}
\end{figure}
}

\newcommand{\exampleSemiFigure}[0]{%
\begin{figure}%
  \centering%
  {%
  \subfigure[%
    Full output of automatic adjudication after
    manually enforcing mentions $(1,3)$ and $(6,6)$
    in chain 1.]{\label{figEnforceIn}%
  \quad%
  \framebox{%
  \BVerbatimInput[numbers=none,firstline=2,lastline=7]{examples_redo1.in.conll}}%
  \qquad%
  }%
  \quad%
  \subfigure[%
    Result of performing automatic adjudication
    relative to enforced coreference information
    in (a).]{\label{figEnforceOut}%
  \quad%
  \framebox{%
  \BVerbatimInput[numbers=none,firstline=2,lastline=7]{examples_redo1.u.conll}}%
  \quad%
  }%
  }%
  \caption{%
  Semi-automatic adjudication of annotations
  given in Figure~\ref{figInFiles}
  using objective \obju.
  Manually enforced coreference information
  in the last column of the CoNLL format
  is prefixed with `\texttt{=}'.
  }%
  \label{figSemiFiles}%
\end{figure}
}
\makeatother

\title{%
{\sc Technical Report:}\\[1em]
Adjudication of Coreference Annotations
via Answer Set Optimization%
}

\author{Peter Schüller%
\thanks{This work extends prior work \citep{Schuller2017lpnmradjudication} with a semi-automatic adjudication encoding, extended formal descriptions and discussions, a tool description, and several additional examples.
This is a preprint of \citep{Schuller2018jetaiadjudication}.
}
\\
Institut f{\"{u}}r Logic and Computation, Knowledge-Based Systems Group\\
Technische Universit{\"{a}}t Wien, Austria\\
Faculty of Engineering, Marmara University, Turkey \\
{\tt schueller.p@gmail.com}
}

\begin{document}
\maketitle
\begin{abstract}
We describe the first automatic approach for merging
coreference annotations obtained from multiple
annotators into a single gold standard. This merging is
subject to certain linguistic hard constraints and
optimization criteria that prefer solutions with
minimal divergence from annotators. The representation
involves an equivalence relation over a
large number of elements.  We use Answer Set Programming 
to describe two representations of the problem
and four objective functions suitable for
different datasets. We provide two structurally different
real-world benchmark datasets based on the METU-Sabanci
Turkish Treebank and we report our experiences in
using the Gringo, Clasp, and Wasp tools for computing
optimal adjudication results on these datasets.
\end{abstract}

\ifdraft
\listoftodos
\fi

\section{Introduction}
\label{secIntro}
Coreference Resolution
\citep{Ng2010coref15,Sapena2008corefsurvey}
is the task of finding phrases in a text
that refer to the same real-world entity.
Coreference is commonly annotated by marking subsequences
of tokens in the input text as \emph{mentions}
and putting sets of mentions into \emph{chains}
such that all mentions in a chain refer to the same, clearly identifiable entity in the world.
\begin{example}[\reva{\citet{Lee2013dcoref}}]
\label{exIntro}
\reva{In the text }%
``\emph{John is a musician.
He played a new song.
A girl was listening to the song.
`It is my favorite,' John said to her.}''
we can identify the following mentions.
\begin{align}
\label{eqExample}
\text{%
\begin{tabular}{@{}p{0.80\textwidth}@{}}
[John]\,$^{(i)}$ is [a musician]\,$^{(ii)}$.
[He]\,$^{(iii)}$ played [a new song]\,$^{(iv)}$. \newline
[A girl]\,$^{(v)}$ was listening to [the song]\,$^{(vi)}$. \newline
``[It]\,$^{(vii)}$ is {\big[}\,[my]\,$^{(ix)}$ favorite{\big]}\,$^{(viii)}$,''
[John]\,$^{(x)}$ said to [her]\,$^{(xi)}$.
\end{tabular}}
\end{align}
Roman superscripts denote mention IDs,
chains in this text are as follows:
$\{(i)$, $(\mi{iii})$, $(\mi{ix})$, $(x)\}$ (John, He, my, John); %
$\{(\mi{iv}), \mi{(vi)}, \mi{(vii)}\}$ (a new song, the song, It); and %
$\{(v), \mi{(xi)}\}$ (A girl, her),
where roman numbers again refer to mention IDs.
\hfill\qed
\end{example}
For building and testing automatic coreference resolution methods,
annotated corpora, i.e.,
texts with mention and chain annotations,
are an important resource.
Once trained, coreference resolution systems can be applied
in various applications,
for example, \citet{Cardie2004} use it for complex opinion extraction,
\citet{Mueller2004} uses it for script-based story understanding,
\citet{Witte2010} perform reasoning on coreference chains to create OWL Ontologies from texts,
and \citet{Tuggener2014} compares the accuracy of coreference resolution systems
when used as preprocessing for discourse analysis, summarization, and finding entity contexts.

Adjudication is the task of combining mention and chain
information from several human annotators
into one single \emph{gold standard} corpus.
These annotations are often mutually conflicting
and resolving these conflicts is a task that is
\emph{global} on the document level,
i.e., it is not possible to decide the truth of the annotation of one token, mention, or chain,
without considering other tokens, mentions, and chains in the same document.

We here present results and experiences obtained
in a two-year project for creating a Turkish coreference
corpus \citep{Schuller2017turkishcorefarxiv},
which included an effort for developing and improving
a (semi)automatic solution for coreference adjudication.
We produced two datasets which are assembled
from~475 individual annotations of~33 distinct documents
from the METU-Sabanci Turkish Treebank~\citep{Say2004}.
{\revamark
Adjudicating documents manually
with tools such as BART~\citep{Versley2008bart}
is usually done on a small set of annotations,
for example, coreference annotations in the OntoNotes
corpus~\citep{Pradhan2007coref} were created by at most
two independent annotators per document.
In the Turkish corpus, we needed to adjudicate
between eight and twelve independent
coreference annotations.
Given such a high number of annotations,
it is suggestive to use majorities of annotator decisions
for suggesting an adjudication solution to the human annotator.

In this paper, we describe a (semi-)automatic solution
for supporting adjudication of
coreference annotations%
}
based on Answer Set Programming (ASP)~%
\citep{Baral2004,Lifschitz2008,Brewka2011cacm,Gebser2012aspbook}.
ASP is a logic programming and knowledge representation paradigm
that allows for a declarative specification of problems
and is suitable for solving large-scale combinatorial
optimization problems.
\smallskip

Our contributions are as follows.
\begin{itemize}
\item
  We formalize the \emph{problem} of coreference adjudication,
  introduce four \emph{objective functions}
  that have practical relevance for both our datasets,
  and we describe the basic idea of
  semi-automatic adjudication
  in Section~\ref{secAutomatic}.
\item
  We propose two \emph{ASP encodings}
  in Section~\ref{secEncodings}:
  the MM (mention-mention) encoding
  explicitly represents a transitive closure
  over the equivalence relation of chains,
  while the CM (chain-mention) encoding
  avoids this explicit representation.
  Moreover, we provide an ASP module
  for semi-automatic adjudication.
\item
  We describe our tool
  and the intended adjudication workflow
  in Section~\ref{secTool}.
\item
  We describe and provide two real-life \emph{datasets},%
  \footnote{\url{https://bitbucket.org/knowlp/asp-coreference-benchmark}}
  outline their properties and differences,
  and report on \emph{empirical experiments}
  with
  unsatisfiable-core optimization and stratification
  using the tools \gringo~\citep{Gebser2011gringo3},
  \clasp~\citep{Gebser2012aij}, and
  \wasp~\citep{Alviano2015wasp}
  in Section~\ref{secEvaluation}.
\item
  We formulate \emph{insights}
  about developing ASP applications,
  analyzing bottlenecks in such applications,
  and specific issues we encountered with optimization
  in Section~\ref{secDiscussion}.
\end{itemize}
We describe related work in Section~\ref{secRelated}
and conclude in Section~\ref{secConclusion}.

Our approach is not specific to Turkish,
and we have implemented it in the publicly available
\crasp\ tool%
\footnote{\url{https://github.com/knowlp/caspr-coreference-tool}}
for performing (semi-)automatic adjudication
of coreference annotations based on the popular
CoNLL data format.
\crasp\ is the first automatic tool
for coreference adjudication,
and our datasets are the first published datasets
for automatic coreference adjudication.

\section{Preliminaries}
\label{secPrelims}

\subsection{Coreference Resolution}
\emph{Coreference resolution}
is the task of finding phrases in a text
that refer to the same entity
\citep{Sapena2008corefsurvey,Ng2010coref15}.
We call such phrases \emph{mentions},
and we call a group of mentions
that refer to one entity a \emph{chain}.
Formally we can describe mention detection
and coreference resolution as follows.

{\revamark
\begin{definition}
  A \emph{document} $D$ is a sequence of tokens $w_1,\ldots,w_n$
  with indexes $1,\ldots,n$.
  A \emph{mention} is a pair $(s,e)$ of indexes,
  $1 \les s \les e \les n$,
  pointing to start index $s$ and end index $e$ in $D$,
  such that the sequence $w_s w_{s+1} \cdots w_e$
  is a natural language expression that refers to a discourse entity
  \citep{Chomsky2002}.
  \emph{Mention detection}
  is the task of finding the set
  $M \eqs \{ (s_1,e_1), \ldots, (s_m,e_m) \}$
  of all mentions in a document.
\end{definition}
\begin{example}\label{exTokensMentions}
  In Example~\ref{exIntro}
  the document~\eqref{eqExample} contains 31 tokens with indexes as follows.
  \begin{align*}
  &\text{%
  John$^1$ is$^2$ a$^3$ musician$^4$ .$^5$
  He$^6$ played$^7$ a$^8$ new$^9$ song$^{10}$ .$^{11}$
  } \\
  &\text{A$^{12}$ girl$^{13}$ was$^{14}$ listening$^{15}$ to$^{16}$ the$^{17}$ song$^{18}$ .$^{19}$} \\
  &\text{``$^{20}$ It$^{21}$ is$^{22}$ my$^{23}$ favorite$^{24}$ ,$^{26}$ ''$^{25}$
  John$^{27}$ said$^{28}$ to$^{29}$ her$^{30}$ .$^{31}$}
  \end{align*}
  The phrases
  \quo{John}, \quo{a musician}, \quo{he}, and \quo{a new song}
  in the first line are mentions,
  and they are represented as pairs $(1,1)$, $(3,4)$, $(6,6)$, and $(8,10)$,
  respectively.
  The overall set of mentions is
  $M \eqs \{ (1,1)$, $(3,4)$, $(6,6)$, $(8,10)$, $(12,13)$, $(17,18)$,
  $(21,21)$, $(23,23)$, $(23,24)$, $(27,27)$, $(30,30) \}$.
  \hfill\qed
\end{example}

\begin{definition}
  Given a set $M$ of mentions,
  \emph{coreference resolution} is the task of partitioning
  $M$ such that for each partition,
  all mentions refer to the same discourse entity.
  The resulting partitions of $M$ are called \emph{chains} or \emph{entities}.
\end{definition}
\begin{example}
  \label{exChains}
  The chains of document \eqref{eqExample}, described in Example~\ref{exIntro},
  are formally represented as
  $P \eqs \{ \{ (1,1)$, $(6,6)$, $(23,23)$, $(27,27) \}$,
  $\{ (12,13)$, $(30,30) \}$,
  $\{ (12,13)$, $(30,30) \} \}$.
  \hfill\qed
\end{example}
}

Mentions can be part of other mentions,
but in that case, they (usually) cannot be coreferent.
Moreover, mentions are (usually) noun phrases \citep{Sapena2008corefsurvey},
which has the consequence that whenever two mentions $m$ and $m'$ are overlapping,
either $m$ is properly contained in $m'$ or vice versa.
{\revamark
Per usual linguistic convention,
mention pairs that are in a predicative relationship
(e.g., \quo{John \emph{is} a musician})
are not coreferent.
In Example~\ref{exIntro},
mention pairs $(\mi{i})$/$(\mi{ii})$
and $(\mi{vii})$/$(\mi{viii})$
are in such a predicative relationship.
This condition cannot be verified automatically
without a structural analysis of the given text,
therefore we do not use it as part of formal constraints
on adjudication solutions.}

Given a set of mentions,
there are exponentially many potential solutions
to the coreference resolution problem,
and finding a globally optimal solution
is NP-hard according to most measures of
coreference optimality
\citep{Stoyanov2012}.

Note that \emph{anaphora resolution}
\citep{Hirst1981,Mitkov1999,Clark2008,Kehler2008}
is a different task than coreference resolution.
The former deals only with references to earlier
parts of a text and sometimes even only with references
where a pronoun points to another phrase.
Contrary to that, coreference resolution
also deals with noun phrases that can refer to each other,
and references can be in any direction within the text.

\subsection{Answer Set Programming}
ASP is a logic programming paradigm
which is suitable for knowledge representation
and finding solutions for computationally (NP-)hard problems~%
\citep{Gelfond1988,Lifschitz2008,Brewka2011cacm}.
We next give brief preliminaries of ASP programs
with (uninterpreted) function symbols,
aggregates, choices, and weak constraints.
For a more elaborate description,
we refer to the ASP-Core-2 standard \citep{Calimeri2012},
or to books about ASP \citep{Baral2004,Gelfond2014aspbook,Gebser2012aspbook}.

\leanparagraph{Syntax}
Let $\cC$ and $\cX$ be mutually disjoint sets
of \emph{constants} and \emph{variables},
which we denote with the first letter in lower case and upper case,
respectively.
Constant names are used for constant terms, predicate names,
and names for uninterpreted functions.
The set of \emph{terms} $\cT$ is recursively defined,
it is the smallest set containing $\bbN \cups \cC \cups \cX$
as well as uninterpreted function terms of form $f(t_1,\ldots,t_n)$ where
$f \ins \cC$ and $t_1,\ldots,t_n \ins \cT$.
An \emph{ordinary atom} is of the form $p(t_1,\dots,t_n)$,
where $p \ins \cC$, $t_1,\dots, t_n \ins \cT$, and
$n\geq 0$ is the \emph{arity} of the atom.
An \emph{aggregate atom} is of the form
$X \eqs \#\mi{agg} \{\ t \cols b_1, \ldots, b_k \ \}$
with variable $X \ins \cX$, aggregation function
$\#\mi{agg} \ins \{ \#\mi{max}, \#\mi{count} \}$,
with $1 \lts k$,
$t \ins \cT$ and $b_1,\ldots,b_k$ a sequence of atoms.
A term or atom is \emph{ground} if it contains no sub-terms
that are variables.

{\revamark
A \emph{rule $r$} is of the form
\begin{equation}
  \label{eqRule}
 \alpha_1 \lors \cdots \lors \alpha_k \leftarrow \beta_1, \dots, \beta_n,
  \naf\, \beta_{n \pluss 1},\dots,\naf\,\beta_{m}
\end{equation}
where $0 \les k$, $0 \les n \les m$,
each head element $\alpha_i$, $0 \les i \les k$, is an ordinary atom,
and each head element $\beta_j$, $0 \les j \les m$, is an atom.
A rule $r$ is a {\em constraint}, if $k \eqs 0$, %
and a \emph{fact} if $k \eqs 1$ and $m \eqs 0$.

A {\em weak constraint} is of form
\begin{equation}
\label{eqWeak}
{\wlar}\, \beta_1, \dots, \beta_n, \naf\, \beta_{n \pluss 1},\dots,
\naf\,\beta_{m}.\ [w@1,t_1,\ldots,t_l]
\end{equation}
where $0 \les n \les m$, $0 \les l$, $w \in \mathbb{Z}$,
$\beta_j$, $0 \les j \les m$, is an atom,
and $t_i$, $0 \les i \les l$, is a term.
(Note that $@1$ determines the `level'
of the weak constraint which we do not use.)

We denote by $H(r) = \{ \alpha_1, \ldots, \alpha_k \}$
the head of a rule $r$.
We denote by $B^+(r) = \{ \beta_1, \dots, \beta_n\}$
the positive body,
by $B^-(r) = \{ \beta_{n \pluss 1},\dots,\beta_m \}$
the negative body,
and by $B(r) = \{ \beta_1, \dots, \beta_n, \naf\ \beta_{n\pluss 1},\dots$, $\naf\ \beta_m \}$
the body of a rule or weak constraint $r$.

A \emph{program} is a finite set $P$ of rules
and weak constraints.

To ensure finite instantiation,
all variables in an ASP rules and weak constraints
must be \emph{safe},
i.e., they must occur in a positive body literal.
For details see the ASP-Core-2 standard \citep{Calimeri2012}.

\leanparagraph{Semantics}
The semantics of an ASP program $P$ is defined
using its Herbrand Base $\HBP$ and its
ground instantiation $\grnd(P)$.
Given an interpretation $I \subseteqs \HBP$,
an atom $a$ is true wrt.\ $I$ iff $a \ins I$
and an aggregate atom
$X \eqs \#\mi{agg} \{\ t \cols b_1, \ldots, b_k \ \}$
is true wrt.\ $I$ iff
$X \eqs f_\mi{\#agg}(\{\ \sigma(t) \mid \sigma(b_1) \ins I, \ldots, \sigma(b_k) \ins I\ \})$
where $\sigma$ is a substitution from variables $\cX$
to nonground terms in $\cT$,
$f_{\mi{\#max}}(X) = \max(X)$ and
$f_{\mi{\#count}}(X) = |X|$.
The body of a rule or weak constraint
is satisfied wrt.\ $I$
iff all $a \ins B^+(r)$ are true wrt.\ $I$
and no $a \ins B^-(r)$ is true wrt.\ $I$.
A rule $r$ is satisfied wrt.\ $I$
iff $H(r) \caps I \eqs \emptyset$
or $B(r)$ is not satisfied wrt.\ $I$.
A weak $r$ constraint is satisfied wrt.\ $I$
iff $B(r)$ is not satisfied wrt.\ $I$.
An interpretation $I$ is a model of a program $P$
if $I$ satisfies all rules in $P$.

The FLP-reduct~\citep{Faber2011} $\fP^I$
reduces a program $P$ using an answer set candidate $I$:
$\fP^I = \{ r \in \grnd(P) \mid B(r)\text{ is satisfied wrt.\ }I \}$.
An interpretation $I$ is an \emph{answer set} of $P$
iff
$I$ is a $\subseteq$-minimal model of $\fP^I$.

Let $W(I)$ be the set of ground weak constraints $r \ins P$
that are not satisfied wrt.\ $I$.
Then the set
$C(I) \eqs \{ (w,t_1,\ldots,t_k) \mid r \in W(I) \}$
is the set of all tuples
of weak constraints of form~\eqref{eqWeak} in $W(I)$,
and the \emph{cost of answer set $I$}
is the sum over the first elements
of all tuples in $C(I)$.
Answer sets of the lowest cost are preferred.}

\leanparagraph{Syntactic Sugar}
Anonymous variables of form \quo{$\_$}
are replaced by new variable symbols.
{\revamark
Choice constructions can occur instead of rule heads
of the form $lb\ \{\ p(\vec{x}) : q(\vec{x})\ \}\ ub$
where lower bound $lb$ and upper bound $ub$ are natural numbers
and can also be omitted (then, $lb \eqs 0$ and $ub \eqs \infty$),
and $\vec{x}$ symbolizes a list of argument terms containing a shared set $x$ of variables.
A rule with a choice head is rewritten into a set of rules with disjunctive heads
of form $p(\vec{x}) \lors \hat{p}(\vec{x})$ for all $q(\vec{x})$,
and into a constraint which enforces that
at least $lb$ and at most $ub$ atoms $p(\vec{x})$ are true
whenever the rule body is true.}
For example,
the rule $1\ \{ p(a) ; p(b) \}\ 2 \leftarrow p(c)$
generates all solution candidates where at least 1 and at most 2 atoms of
the set $\{p(a),p(b)\}$ are true whenever $p(c)$ is true.
{\revamark
For details of choice constructions
and the rewriting to disjunctive rules plus constraints with aggregates,
we refer to Section~3.2 of the ASP-Core-2 standard \citep{Calimeri2012}.}

\section{Automatic Coreference Adjudication}
\label{secAutomatic}
{\revamark
Coreference adjudication is the task of arbitrating several coreference resolution solutions
of a single document
that were independently created by several human annotators.
We assume that human annotators created these solutions
to the best of their knowledge and abilities
and we assume that humans sometimes make mistakes.
Therefore, we are dealing with a problem of merging multiple,
potentially mutually inconsistent, solutions
of the mention detection and the coreference resolution problem.
The goal is to create a single solution that optimally corresponds to the given inputs.

We formalize coreference adjudication }as follows.
\begin{definition}\label{defAdjudication}
  Given $u \ges 2$ sets of mentions $M_1,\ldots,M_u$ over a document $D$
  and corresponding chains $P_1,\ldots,P_u$ of the respective sets of mentions,
  {\revamark
  the \emph{coreference adjudication problem} is to find
  the set of mentions $\hat{M}$ and the set of chains $\hat{P}$
  (with $\hat{P}$ a partition of $\hat{M}$)
  such that $\hat{P}$ and $\hat{M}$ observe structural constraints
  described in Section~\ref{secPrelims},
  and that resulting chains $\hat{P}$ correspond best
  with the given chains $P_1,\ldots,P_u$.}
\end{definition}
{\revamark
The problem of deciding what \emph{corresponds best} is usually solved manually
by domain experts, and usually there is only a small set of given annotations
which mostly coincide because they were created by trained experts.
In our work on the Turkish corpus \citep{Schuller2017turkishcorefarxiv}
we encountered the problem of adjudicating a larger set of annotations
(at least eight annotations per document)
with a higher level of divergence among (less trained) annotators.

We approach this problem by representing a given coreference chain
in terms of all links between pairs of mentions in that chain (mention-mention links).
We represent adjudication by selecting a subset of the given mention-mention links
and constructing a new set of chains from these selected links.
We choose the optimal solution by
defining quantitatively how \quo{correspond best}
in Definition~\ref{defAdjudication} can be measured in terms of selected and non-selected links.}

Note that a solution might also contain mention-mention links
that are not present in any annotation.
This can happen because chains are equivalence relations:
if we merge equivalence relations that are not
subsets of each other, the new equivalence relation
is the reflexive, symmetric, and transitive closure
of the original relations.

\begin{example}
\label{exTransitivityViolation}
Assume that the chains in Example~\ref{exIntro}
\reva{are given by annotator A,
and that annotator B provides }the chain
$\mi{\{(i), (ii), (x)\}}$ (John, a musician, John).
If we merge \reva{ this chain
with the chain $\mi{\{(i), (iii), (ix), (x)\}}$
(John, He, my, John) from annotator A
in a naive way (i.e., by merging the sets),
then }we obtain a single chain
$\mi{\{(i), (ii), (iii), (ix), (x)\}}$
(John, a musician, He, my, John)
although no annotator indicated that the mentions
$\mi{(ii)}$ %
and $\mi{(iii)}$ %
(`a musician' and `He')
belong to the same chain.
\hfill\qed
\end{example}

Next, we give a symbolic and more comprehensive
running example.
\begin{example}\label{exIn1234}
Figure~\ref{figInFiles} shows
an adjudication problem consisting of input from
four annotators.
We show token index in the first column
and coreference chains in the second column
of the CoNLL format,
where single-token mentions are marked as `(X)',
multi-token mentions start with `(X' and end with `X)',
{\revamark
where X is the chain that the mention belongs to.
The absence of any mention starting or ending at a token is marked with `-'.}

In this example,
{\revamark
annotator (b) indicated a single chain with ID~3
containing mentions $\{(1,3),(5,5)\}$,}
and annotator (a) indicated
a chain with ID~2 containing mentions $\{(1,3),(6,6)\}$
as well as a chain with ID~3 containing mentions
$\{(2,2),(4,4)\}$.

There are several possible outcomes
for merging these (input) annotations
into a single (output) gold standard annotation.
Intuitively, annotators (a) and (c) indicated
that mention $(1,3)$ is coreferent with $(6,6)$,
while annotators (b) and (d) indicated
that $(1,3)$ is coreferent with $(5,5)$.
Only annotator (c) marked $(1,1)$ as a separate mention
that is coreferent with $(4,4)$.
\hfill\qed
\end{example}
\exampleInputsFigure%

{\revamark
We next define the meaning of \quo{correspond best}
in Definition~\ref{defAdjudication}
by defining an objective function}
for selecting preferred solutions.

\subsection{Objective Functions}
\label{secObjectives}
We next represent \reva{the given annotations $P_1,\ldots,P_u$} and
\reva{the solution chain $\hat{P}$}
as sets of links between pairs of mentions
(men\-tion-mention links).
We define a preference relation
for filtering out undesired solutions
where we incur cost under the following conditions.
\begin{enumerate}
\item[(C1)]
  Not using a link provided by an annotator in the solution.
\item[(C2)]
  Using a link provided by an annotator
  where a number of other annotators
  did not provide the same link.
\item[(C3)]
  Putting two mentions into a chain
  where no annotator gave any evidence for this
  (see Example~\ref{exTransitivityViolation}).
\end{enumerate}
We incur separate cost for each annotator who provided a link in (C1),
and for each annotator who did not provide a link that was used in (C2).

Concretely we use the following objective functions in our application:
\begin{itemize}
\item[\hphantom{\tt{}U}\objv] Cost 2 for (C1), cost 1 for (C2), and (C3) is a hard constraint (infinite cost).
\item[\hphantom{\tt{}U}\obju] Cost 2 for (C1), no cost for (C2), and (C3) is a hard constraint.
\item[\objva] Cost 2 for (C1), cost 1 for (C2), and cost 1 for (C3).
\item[\objua] Cost 2 for (C1), no cost for (C2), and cost 1 for (C3).
\end{itemize}
Note that cost for (C1) is higher than for (C2) because
we observed \reva{in an initial manual adjudication of two documents of the Turkish corpus}
that annotators miss links more frequently
than they add spurious links.
{\revamark
In~\obju\ only one cost component (C1) is active,
therefore the absolute value of the cost incurred for each link does not matter.
We use the value of~2 in~\obju\ because this makes the coefficient of (C1)
uniform across all objective variations and permits a modular
ASP translation of all objective cost components.}

Intuitively, objectives containing letter {\tt{}V}
apply ``voting'' to prefer certain solutions:
a link given by only one annotator can be
dismissed if many other annotators do not give the same link.
On the other hand, objectives with letter {\tt{}U}
use ``as many mentions as possible'':
there is no cost for (C2).
Finally, objectives containing letter {\tt{}A}
allow additional links at a cost.
{\revamark
In our real-world application of the Turkish coreference corpus,
we opted to use objective \objv\ where (C3) is a strict constraint.
On our data, using \objva\ and choosing a higher value than 1 for the cost of (C3)
had a similar effect as using a strict constraint.
This is
because a solution that merges two chains such that a cost of type (C3) is incurred
usually incurs also several costs of type (C2) in other links.
For other adjudication projects which use objectives \objua\ or \objva,
the coefficient of cost type (C3) might need to be adjusted.}

{\revamark
Cost components are optimized such that their sum is minimized.
There is no lexicographic order (priority) among costs
where one type of cost would be minimized before another type of cost.
For example, the objective function \objv\
finds a balance between ignoring and using annotations that
were given by some and not given by other annotators.
Using a priority would strictly prefer one over the other.%
}

\exampleOutputsFigureLastColumn%
\begin{example}
  \label{exAdjudication}%
  Figure~\ref{figOutFiles} shows automatically adjudicated
  results for all objective functions,
  based on
  \reva{Example~\ref{exIn1234} and Figure~\ref{figInFiles}.}

  Objective \obju\ in Figure~\ref{figOutu}
  produces a single optimal adjudication solution
  where both chains from Figure~\ref{figIn1} and
  chain~1 from Figure~\ref{figIn3} are ignored,
  which yields a total cost of~6
  and ignores three mention-mention links.

  Objective \objua\ in Figure~\ref{figOutua}
  produces a single optimal
  adjudication solution of cost~4:
  chain~4 from Figure~\ref{figIn4} is ignored (cost~2)
  and two links
  that are not present in any annotation
  are created (each incurs cost~1):
  between~$(1,1)$ and~$(2,2)$,
  and between~$(5,5)$ and~$(6,6)$.

  Objective \objv\ produces two optimal
  adjudication solutions, each with cost~12.
  The left solution in Figure~\ref{figOutv}
  uses chain~2 from Figure~\ref{figIn1},
  which is equivalent to chain~1 in Figure~\ref{figIn3}.
  This incurs a cost of~2 because links of this chain
  are absent in Figures~\ref{figIn2} and~\ref{figIn4}.
  All other chains
  (one each in Figures~\ref{figIn1}, \ref{figIn2}, and~\ref{figIn3},
  and two in Figure~\ref{figIn4})
  are ignored at a total cost of~10
  (each of these chains contains one link).
  Using any of these ignored chains
  would incur a cost of~3 per chain
  because the chain was not given
  by three out of four annotators.
  Using chain~3 in Figure~\ref{figIn2} or chain~2 in Figure~\ref{figIn4}
  would create an additional link between~$(5,5)$ and~$(6,6)$,
  which is not permitted in objective~\objv.
  The second solution (Figure~\ref{figOutv} right)
  is symmetric to the first one:
  it uses only chain~3 from Figure~\ref{figIn2},
  which is equivalent to chain~2 in Figure~\ref{figIn4}.

  Objective \objva\ produces a single optimal
  solution of cost~11 that is shown in Figure~\ref{figOutva}.
  Intuitively,
  this solution merges both solutions of \objv.
  Using chain~2 from Figure~\ref{figIn1},
  chain~3 from Figure~\ref{figIn2},
  chain~1 from Figure~\ref{figIn3},
  and chain~2 from Figure~\ref{figIn4}
  incurs cost~4 because each link
  is not annotated by two out of four annotators.
  Moreover, three links
  (chains~3 in Figure~\ref{figIn1},
  chain~2 in Figure~\ref{figIn3},
  and chain~4 in Figure~\ref{figIn4})
  are ignored, which contributes an additional cost of~6.
  The link between~$(5,5)$ and~$(6,6)$,
  which was not specified by any annotator,
  incurs a cost of~1.
  \hfill\qed
\end{example}

The main idea of these preferences is
to use the given information optimally
while producing an overall consistent solution.
The preferences are motivated by properties of our datasets:
if mentions are given to annotators,
they can only disagree on the assignment of mentions to chains.
Objectives \objv\ and \objva\ make sure that the solution reflects
the opinion of the majority of annotators.
Contrarily, if mentions are not given,
annotators must produce mentions and chains,
and often disagree on mentions,
such that \objv\ and \objva\ eliminate most mentions and chains completely.
In the latter case, \obju\ and~\objua\ combined with
a semi-automatic workflow is the better choice.

\subsection{Semi-automatic Adjudication}
Automatic adjudication works well if enough annotations
with high inter-annotator agreement \citep{Passonneau2004} are available.
Otherwise, automatic adjudication is merely a
preprocessing step for human adjudication.
In case of a low number of annotations
per document, or highly divergent annotator opinions,
making the final decision on the correctness of an annotation
should be done by a human expert.

For that purpose, we allow a partial specification
of chain and mention information as additional input.
We extend our approach so that it produces
an optimal solution relative to this given information.

In practice, we do this by producing a human-readable
output format in automatic adjudication.
This format is in the popular CoNLL format
(see Figure~\ref{figSemiFiles})
and contains tokens
and coreference information from all annotators
and from the automatic adjudication.
The idea is that a human adjudicator can inspect the output
and manually specify parts of the output,
followed by re-optimization of the adjudication
relative to these manually enforced parts of the output
(more details about this is given in Section~\ref{secTool}).
\exampleSemiFigure%
\begin{example}
\label{exSemiConll}
Figure~\ref{figSemiFiles} shows a CoNLL file
that was generated from automatic adjudication
of annotations in Figure~\ref{figInFiles}
using objective \obju.
The rightmost column originally was the same as in
Figure~\ref{figOutu} but the human adjudicator
has specified the following coreference information
(prefixed with \quo{\texttt{=}}):
mentions $(1,3)$ and $(6,6)$ must be in a chain,
and mention $(1,1)$,
which was coreferent with $(4,4)$, does not exist.

Re-optimizing the input in Figure~\ref{figEnforceIn}
yields the output Figure~\ref{figEnforceOut}
where the result of automatically adjudicating
relative to enforced mention and chain information
is put into the last column.
Different from the adjudication result explained in
Example~\ref{exAdjudication},
coreference between~$(1,3)$ and~$(5,5)$
is no longer possible because it would generate a
non-annotated link between~$(5,5)$ and~$(6,6)$.
Therefore token~5 carries no mention information
(`{\tt{}-}').
Moreover, the link between~$(1,1)$ and~$(4,4)$
is ruled out because mention~$(1,1)$
was enforced to be absent,
which makes the link between~$(2,2)$ and~$(4,4)$
appear in the optimal solution.
\hfill\qed
\end{example}

\section{ASP Encodings}%
\label{secEncodings}%
We next describe ASP input and output representations
for the adjudication problem.
Then we provide two ASP encodings
which model coreference adjudication and
assign a cost to solutions according to Section~\ref{secObjectives}.
The MM Encoding explicitly represents the transitive closure
of mention-mention links,
while the CM encoding avoids this representation by
assuming an upper limit  on the number of chains
and guessing which mention belongs to which chain.

In all encodings,
we use the convention that variables
$C$, $A$, $\mi{Am}$, $S$, and $E$,
will be substituted with
chain IDs, annotator IDs, annotator mention IDs,
start token indexes, and end token indexes, respectively.
{\revamark
Moreover, variable $M$ holds terms of the form $mid(S,E)$
for representing a mention without annotator information.
We call mentions that are represented in this way
\emph{canonical mentions} or short \emph{cmentions}.}
We use subscripted versions of the variables introduced above.

\subsection{Input and Output ASP Representation}
\label{secInputOutputRepresentation}
{\revamark
Given a document $D$
and $u$ coreference annotations $P_1,\ldots,P_u$
which are partitions of sets of mentions $M_1,\ldots,M_u$ over $D$.
We associate unique IDs $1,\ldots,u$ with annotations.
For each annotation $P_a$, $1 \les a \les u$
we have a set of chains $P_a = \{ C_1, \ldots, C_k\}$
and we associate unique IDs $1,\ldots,k$ with these chains.
The corresponding mentions are $M_a = \{ (s_1,e_1), \ldots, (s_m,e_m) \}$
and we associate unique IDs $1,\ldots,m$ with the mentions in $M_a$.
(In the general case, each annotator produces her own set of mentions.)

\begin{definition}
To represent an \emph{adjudication instance} in ASP,
we create facts of the form $\mi{mention(a,j,s_j,e_j)}$
for each mention $(s_j,e_j)$ with ID $j$ in the annotation with ID $a$.
Moreover, we create facts of the form $\mi{cm(a,c,j)}$
for each mention with ID $j$
that is in the chain with ID $c$
in the annotation with ID $a$.
\end{definition}
}
\begin{example}
  \label{exASPRep}
  Given the document in Example~\ref{exIntro} and token indexes
  shown in Example~\ref{exTokensMentions}.
  Consider that annotator $a_1$ created
  {\revamark the chain $c_1 \eqs \{m_1, m_2, m_3\}$ containing mentions
  $m_1 \eqs (1,1)$ \quo{John},
  $m_2 \eqs (3,4)$ \quo{a musician}, and
  $m_3 \eqs (6,6)$ \quo{He}.
  Annotator $a_2$ created the chain
  $c_2 \eqs \{m_4, m_5\}$
  containing mentions
  $m_4 \eqs (12,13)$ \quo{a girl}
  and $m_5 \eqs (30,30)$ \quo{her}.
  These annotations are represented in ASP as follows:
  \begin{align*}
  &\mi{mention(a_1,m_1,1,1).} & &\mi{mention(a_1,m_2,3,4).} & &\mi{mention(a_1,m_3,6,6).} \\
  &\mi{cm(a_1,c_1,m_1).} & &\mi{cm(a_1,c_1,m_2).} & &\mi{cm(a_1,c_1,m_3).} \\
  &\mi{mention(a_2,m_4,12,13).} & &\mi{mention(a_2,m_5,30,30).} \\
  &\mi{cm(a_2,c_2,m_4).} & &\mi{cm(a_2,c_2,m_5).}
  \end{align*}
  where we use constants $a_1,a_2,c_1,c_2,m_1,\ldots,m_5$ as IDs
  for representing the corresponding annotators, chains, and mentions.
  }
\hfill\qed
\end{example}

The answer sets of our logic program
represent a set of chains without annotator information,
represented as atoms of the form
$\mi{result_{cm}(C,mid(S,E))}$,
which indicates that in chain $\ilv{C}$
there is a mention that is a span from
start token $\ilv{S}$ to end token $\ilv{E}$, inclusive.

\begin{example}
Assume we have adjudicated the annotations
\reva{of Example~\ref{exASPRep}
such that the solution comprises }%
two chains $v \eqs \{ (1,1), (6,6) \}$
and $w \eqs \{ (12,13), (30,30) \}$.
This is represented by the following atoms:
\begin{align*}
&\mi{result_{cm}(v,mid(1,1))} & &\mi{result_{cm}(v,mid(6,6))} \\
&\mi{result_{cm}(w,mid(12,13))} & &\mi{result_{cm}(w,mid(30,30))}
\end{align*}
Intuitively,
this means that `John' and `He' are coreferent,
as well as `a girl' and `her',
however `John' is not coreferent with `a musician'.
\hfill\qed
\end{example}%

Input for \emph{semi-automatic adjudication}
consists of mentions, chains,
and \reva{indicators for the absence of mention boundaries}.
Mentions and chains are represented as
a separate annotation with a special annotator ID
\quo{$\mi{forced}$}.
A token $\mi{Token}$ without any mention boundary is represented as
a fact of form $\mi{empty(Token)}$.
\begin{example}
In \reva{Example~\ref{exSemiConll}
(Figure~\ref{figEnforceIn}) }%
the human adjudicator
specified that mention $(1,3)$ is coreferent with $(6,6)$.
This is represented by the following facts.
\begin{align*}
  &\mi{mention(forced,m_1,1,3) \lars.} &
  &\mi{mention(forced,m_2,6,6) \lars.} \\
  &\mi{cm(forced,c_1,m_1) \lars.} &
  &\mi{cm(forced,c_1,m_2) \lars.}
\end{align*}
where the chain is represented by constant $c_1$
and the mentions by $m_1$ and $m_2$.
\hfill\qed
\end{example}

We next describe our encodings.

\myencodingfigurecostcommonMath%
\subsection{Common Rules for Objective Costs}
\label{secCommonEncodingsEvidence}
Figure~\ref{figCommon} shows a common program module
which defines helpful concepts for realizing
cost aspects (C1)--(C3) from Section~\ref{secObjectives}.
The rules
{\revamark contain no disjunctions or guesses}
and depend only on input facts,
hence this module is deterministic
{\revamark and a modern grounder like \gringo\ will transform it into facts during instantiation}.

Rule~\eqref{eqNofAnno} represents the number of
annotations in the input in \reva{predicate }$\mi{na}/1$.
This number is used for relating the weight
of one annotator's input compared with all annotators.
Rule~\eqref{eqAcmen} defines predicate $\mi{acmen}/3$
which relates annotated mentions
with their canonical representation in a term of the form
$\mi{mid(S,E)}$ where $S$ and $E$ are the starting and
ending token of the mention in the document\reva{, respectively}.
Rule~\eqref{eqASameChain} represents in
{\revamark atoms of the form
$\mi{asamechain}(A,\mi{Am}_1,\mi{Am}_2)$
all pairs of distinct mentions $\mi{Am}_1$ and $\mi{Am}_2$
that were put into the same chain by annotator $A$.}
Rule~\eqref{eqCSameChain} represents the same
{\revamark pairs of mentions
but projects away annotator information
and relates pairs of cmentions in atoms of the form $\mi{csamechain}(M_1,M_2)$.}

Based on $\mi{csamechain}$ and $\mi{asamechain}$,
rule~\eqref{eqPosEvidence} represents
for each pair of canonical mentions $(M_1,M_2)$
{\revamark
the number of annotators who put $M_1$ and $M_2$ into the same chain.}
This value is represented only if it is above zero.
Rule~\eqref{eqOmitCost} defines cost component (C1)
which is a cost of 2 for each annotator who provided
positive evidence for the respective link.
Similarly, using the total number of annotators,
rule~\eqref{eqUseCost} defines cost component (C2)
which is a cost of 1 for each annotator
who did not put both canonical mentions $M_1$ and $M_2$
into the same chain,
while at least one other annotator did so.

\myencodingfigurecommonLinkMath
{\revamark
\subsection{Common Link Guess and Canonicalization Encoding}
\label{secCommonEncodingsLinkGuess}
Figure~\ref{figCommonLink} shows a nondeterministic common module
of both encodings.
This module guesses for each mention-mention link
that was annotated whether it will be part of the solution.
Selected mention-mention links are represented canonically in predicate $\mi{clink}/2$.

In detail, }%
Rule~\eqref{eqLinkFromACM} represents annotated
mention-mention links in predicate $\mi{link}/3$,
and rule~\eqref{eqGuessUseLink} is a guess
whether to use each of these links.
Used links are represented as cmentions
in rules~\eqref{eqCanonicalize1}
and~\eqref{eqCanonicalize2},
i.e., their annotator information is projected away.%
\footnote{This \reva{projection }requires two rules
because mention and cmention IDs
can have different lexicographic order.}
Note that these rules make use of deterministically defined
predicate $\mi{acmen}/3$ from the common program module
given in Figure~\ref{figCommon}.

\myencodingfiguremmMath%
\subsection{Mention-Mention Encoding (MM)}
Figure~\ref{figMM} shows the core module of the MM encoding
{\revamark
which is used in addition to the common program modules in Figures~\ref{figCommon} and~\ref{figCommonLink}.}
Rules~\eqref{eqMMCCCopy}--\eqref{eqMMCCTransitive} represent the reflexive, symmetric, and transitive closure
of \reva{predicate }$\mi{clink/2}$ in \reva{predicate }$\mi{cc/2}$.
This represents result chains in an explicit equivalence relation over cmentions:
each strongly connected component (SCC) of \reva{predicate }$\mi{cc/2}$
is equivalent to a coreference chain.
Rule~\eqref{eqMMNotRep} defines which elements
of the equivalence relation is not the lexicographically
smallest element in the SCC,
and rule~\eqref{eqMMResultCM} defines result chains
by using the smallest cmention in each SCC
as a representative for the corresponding chain.
Rule~\eqref{eqMMResultC} defines that these cmentions represent chains.

Figure~\ref{figMMObjective} shows constraints
for realizing objective functions \reva{for the MM encoding}.
Weak constraint~\eqref{eqMMOmitCost} incurs a cost
for non-existing links in \reva{atoms of form }$\mi{clink}(M_1,M_2)$,
corresponding to cost~(C1) in Section~\ref{secObjectives}.
Similarly,~\eqref{eqMMUseCost} incurs cost for existing links,
corresponding to~(C2).
Finally, corresponding to~(C3),
constraint~\eqref{eqMMTransForbid} strictly forbids to put two mentions into a chain if there is no
evidence for that from annotators,
and weak constraint~\eqref{eqMMTransCost}
alternatively incurs a cost for such mention pairs.
{\revamark
Recall that predicates $\mi{cmomitcost}/3$ and $\mi{cmusecost}/3$
are defined deterministically from input facts
(see Section~\ref{secCommonEncodingsEvidence} and Figure~\ref{figCommon}),
therefore weak constraints~\eqref{eqMMOmitCost} and~\eqref{eqMMUseCost}
contain only one body literal after intelligent instantiation.}

\myencodingfigurecmMath%
\subsection{Chain-Mention Encoding (CM)}
Figure~\ref{figCM} shows the core of the CM encoding,
{\revamark
which is also based on the common program modules in Figures~\ref{figCommon} and~\ref{figCommonLink}.
Different from the MM encoding,
in the CM encoding we directly guess the extension of predicate }$\mi{resultcm/2}$
instead of deriving it from \reva{predicate }$\mi{clink/2}$.

{\revamark
Rules~\eqref{eqCMCountChain}--\eqref{eqCMChainLimit}
deterministically depend on input facts.}
Rule~\eqref{eqCMCountChain} counts the number of chains
per annotator, \eqref{eqCMMaxChain} \reva{represents }the maximum
number of chains that any of the annotators created,
and~\eqref{eqCMChainLimit} multiplies this number
with $6/5$.
This value serves as an assumption for the maximum number \reva{$N$}
of chains in the result
and deterministically depends on the given instance.
We found that \reva{this }assumption %
can safely be used in practice
because annotators are consistent in the number
of \reva{chains }they produce over the whole document,
i.e., annotators either create many chains or few chains
relative to other annotators,
but they never create many chains in some part of the document
and few chains in other parts of the same document.
(Here, many and few are relative to other annotators.)

{\revamark
In rule~\eqref{eqCMResultChain}
we nondeterministically guess which chains of the assumed maximum of $N$ chains
actually exists in the result.}
Rule~\eqref{eqCMResultChainSB} performs symmetry breaking
on this guess to use a gap-free sequence of chain IDs
starting at 1.

Rules~\eqref{eqCMCm1} and~\eqref{eqCMCm2}
represent \reva{all }cmentions that
appear in canonical links \reva{which are represented in $\mi{clink}/2$.
(Note that these are defined in the module in Figure~\ref{figCommonLink}.)}
In rule~\eqref{eqCMResultChainMention}
we guess which cmention is contained in which result chain.
Constraints~\eqref{eqCMSync1} and~\eqref{eqCMSync2}
ensure that the links \reva{represented }in $\mi{clink/2}$
which were selected by \reva{the guess in rule}~\eqref{eqGuessUseLink}
are fully represented by \reva{predicate }$\mi{resultcm/2}$.

\myencodingfigurecmobjMath
Figure~\ref{figCMObjective} shows objective
functions for CM.
Weak constraints~\eqref{eqCMOmitCost} and~\eqref{eqCMUseCost}
for cost components (C1) and (C2)
are the same as~\eqref{eqMMOmitCost} and~\eqref{eqMMUseCost}
in the MM encoding.
Constraints for realizing (C3) are different
because in the CM encoding
we have no transitive closure $\mi{cc}/2$ at our disposal.
Therefore, in constraint~\eqref{eqCMTransForbid}
we need to make an explicit join over $\mi{resultcm}/2$
to rule out mention-mention links in the result
that have no corresponding link in $\mi{clink}/2$.
Weak constraints~\eqref{eqCMTransCost}
alternatively incur a cost for such links.
\smallskip

\mytableObjectiveConstraints%
Table~\ref{tblObjectiveConstraints}
shows the combinations of constraints that realize
each objective function.
Intuitively, we always incur a cost for \reva{not using annotated mention-mention links }via~\eqref{eqMMOmitCost},
\reva{in objectives containing {\tt{}V} }we additionally incur a cost for
\reva{using mention-mention links that were not annotated by all annotators }via~\eqref{eqMMUseCost}.
Those mention-mention links that were not annotated at all
\reva{ but appear in the solution
(due to building a transitive closure over links annotated by multiple annotators)}
we either forbid them completely via~\eqref{eqMMTransForbid}
or we incur an additional cost via~\eqref{eqMMTransCost}.

{\revamark
\subsection{Module for Common Structural Constraints}
Figure~\ref{figCommonConstraints} shows an encoding module
for common structural constraints of adjudication solutions.

In most coreference corpora,
singleton chains, i.e., chains that contain only a single mention,
are not permitted.
For that purpose,
constraint~\eqref{eqNoSingletons} enforces
that each chain contains at least two mentions.

Moreover, it is linguistically motivated
that a mention cannot corefer to another mention
that is a sub-phrase of the former mention.
Constraint~\eqref{eqNoNestedCoref} requires
that no chain contains two mentions
where one is contained in the other.}

\subsection{Semi-automatic Encoding Module}
So far we have covered only automatic adjudication.
Semi-automatic adjudication is based on
enforced mention and chain information,
and on tokens that are enforced to be empty
as described in Section~%
\ref{secInputOutputRepresentation}.

\myencodingfigureCommonConstraints
\myencodingfigureenforceMath%
Figure~\ref{figEnforceMath} contains an encoding module
for semi-automatic adjudication
which can be used in addition to the CM or MM encodings
and in combination with any of the objective functions.
To ensure that annotations with special annotator ID
`\mi{forced}' are reproduced in every solution,
we first represent several auxiliary concepts.
Rules~\eqref{eqForceEmpty}--\eqref{eqForceEnd}
represent which tokens contain (en-)forced annotations,
rule~\eqref{eqForceSame} represents pairs of
canonical mentions that must be part of the same chain,
and rule~\eqref{eqForceDiff} represents pairs
that must be part of different chains.
Constraint~\eqref{eqForceNoInventS} ensures
that no canonical mention starts at an enforced token
if that mention was not enforced,
and~\eqref{eqForceNoInventE} ensures the same
for ends of mentions.
Note that these two constraints and~\eqref{eqForceEmpty} ensure
that enforced empty tokens obtain no mention annotations.

Constraint~\eqref{eqForceReproduce} ensures
that all enforced mentions exist in the solution
as canonical mentions.
Finally, constraint~\eqref{eqForceSameCheck}
ensures that mentions which should be in the same chain
are not in different chains,
and constraint~\eqref{eqForceDiffCheck}
ensures that mentions which should be in distinct chains
are not in the same chain.

Adding these rules to encoding MM or CM is sufficient
for ensuring that only solutions
that reproduce the user-specified annotations
remain as answer sets.

\section{Tool and Adjudication Workflow}
\label{secTool}
We have implemented our method in an open source tool
called \crasp\ which is available publicly at
\url{https://github.com/knowlp/caspr-coreference-tool}
{\revamark
and implements encodings MM
(Figures~\ref{figCommon}, \ref{figCommonLink}, and~\ref{figMM})
and CM
(Figures~\ref{figCommon}, \ref{figCommonLink}, and~\ref{figCM}).
For both encodings,
all four objective functions \obju, \objua, \objv, and \objva\
(Figures~\ref{figMMObjective} and~\ref{figCMObjective})
can be selected.
Structural constraints (Figure~\ref{figCommonConstraints})
are activated per default.
Semi-automatic mode (Figure~\ref{figEnforceMath})
can be activated on the command line.}

\crasp\ reads multiple files in CoNLL format
(similar to our examples, potentially with additional columns),
where each file contains annotations from one annotator.
After automatic adjudication,
a new CoNLL file is produced that contains either only
the resulting annotation or all input annotations and the
adjudication solution in separate columns of a single file
for manual inspection.
The result column of this file can be edited
in order to enforce mentions or absence thereof,
and \crasp\ can reprocess this file in semiautomatic mode.
\crasp\ is realized based on
\gringo\ \citep{Gebser2011gringo3} and
\clasp\ \citep{Gebser2012aij} version 5.

\smallskip

Our tool facilitates the following semiautomatic
coreference adjudication process.
\begin{enumerate}
\item
  Use \crasp\ to obtain a consistent adjudication of annotations
  with optimal usage of given information.
\item
  Review the resulting CoNLL file in an editor,
  and enforce coreference information of certain tokens
  if they are clearly wrong
  according to the expertise of the human adjudicator.
  Coreference information at a token is enforced
  by prefixing it with the character `{\tt{}=}'.
\item
  Run the re-adjudication mode of \crasp,
  which creates an optimal adjudication
  relative to manually enforced tokens.
\item
  Iterate the steps (2) and (3) until a satisfactory
  gold standard result is obtained.
\end{enumerate}
Importantly, step (3) only changes tokens
that have not been enforced,
hence an incremental workflow,
starting with the most clear-cut cases,
and ending with the most difficult cases,
can be followed.

The CoNLL data format used in \crasp\ is variation
used by the
CorScorer
\citep{Pradhan2014scorer}
reference coreference resolution scoring tool.
Similar to CorScorer, \crasp\ expects
\texttt{\#begin document} and \texttt{\#end document} tags (which have been omitted in figures for brevity),
considers the last column to be the coreference annotation column,
and silently copies all other columns into the result.

For our purposes, it was sufficient to use \crasp\
directly on CoNLL files and to use a text editor as GUI.
To make the tool accessible to a wider part
of the community,
we consider integrating it into an existing
coreference annotation toolkit,
for example into BART~\citep{Versley2008bart}.

\section{Evaluation}
\label{secEvaluation}
We first describe our datasets and then the experimental results.
\subsection{Datasets}
\label{secDataset}

The datasets that prompted development of this application
are based on the METU-Sabanci Turkish Treebank \citep{Say2004,Atalay2003,Oflazer2003turkishtreebank}.
Table~\ref{tblDatasets} shows the properties of both datasets.
Documents were annotated in two distinct annotation cycles:
for DS1, annotators had to produce mentions and chains,
while for DS2, mentions were given and could be assigned to chains or removed by annotators.
This yielded a large \reva{number }of \reva{distinct }mentions,
\reva{i.e., mentions over all annotators with distinct mention boundaries}
for DS1 (on average~316.7 per document),
while DS2 contains fewer \reva{distinct }mentions~(159.8)
although it contains by far more \reva{annotated }mentions~(1561.4)
than DS1.
DS1 is also smaller; it is based on~21 documents
from the corpus while DS2 covers the whole corpus \reva{of~33 documents}
and has more annotations per document.
\reva{The average document size is~1634 tokens per document.}

In practice we observed the following:
due to disagreement on canonical mention boundaries in DS1,
adjudication with~\objv\ or~\objva\ eliminates most data,
hence using~\obju\ or~\objua\ is more useful.
On the other hand, the `voting' of~\objv\ and~\objva\
can utilize the larger number of annotations per canonical mention in DS1,
which yields practically very usable results
that do not require manual intervention for creating
a gold standard.
DS1 and DS2 are structurally quite different:
DS1 has several instances where nearly all mentions
are (transitively) connected to all other mentions,
while this does not occur in DS2.

\mydatasettable%
\subsection{Experiments}
\label{secExperiments}
Experiments were performed on a computer with~48~GB RAM
and two Intel E5-2630 CPUs (total~16 cores)
using Debian~8 and at most seven concurrently running jobs,
each job using at most two cores
(we did not use tools in multi-threaded mode).
As systems, we used \gringo\ and \clasp\
from \clingo~5 (git hash~933fce)~%
\citep{Gebser2011gringo3,Gebser2012aij}
and \wasp~2 (git hash~ec8857)~\citep{Alviano2015wasp}.
We use \clasp\ with parameter~{\tt\small{}-{}-opt-strategy=usc,9}
and \wasp\ with parameter~{\tt\small{}-{}-enable-disjcores},
i.e., both systems use unsat-core based optimization
with stratification and disjoint core preprocessing.
\reva{We always activate unsat-core based optimization
because branch-and-bound~({\tt\small{}-{}-opt-strategy=bb} for \clasp\
and~{\tt\small{}-{}-weakconstraints-algorithm=basic} for \wasp)
yielded timeouts for nearly all instances in preliminary experiments.
We enabled disjunctive cores for unsat-core optimization
because it slightly increased performance in preliminary experiments.}

Table~\ref{tblExp} shows experimental results.
We limited memory usage to 5~GB and time usage
to 300~sec and we show results
averaged over 5 repeated runs for each configuration.
Columns MO, respectively TO, show the number of runs
that aborted because of exceeding
the memory limit (memory out),
respectively the time limit (timeout).
Columns SAT and OPT show the number of runs
that found some solution and the first optimal solution,
respectively,
moreover, we show the percentage of runs
that found an optimal solution.
Columns $T$ and $M$ give average time and memory usage
which was measured with the \runlim\ tool.%
\footnote{\url{http://fmv.jku.at/runlim/}}
Columns $T_\mi{grd}$, $\mi{Opt}$, $\mi{Chc}$, and $\mi{Cnf}$
show average instantiation time,
optimality of the solution ($\frac{\mi{UB}{-}\mi{LB}}{\mi{LB}}$), number of choices,
and number of conflicts encountered in the solver.
To make values comparable,
columns $T_\mi{grd}$ through $\mi{Cnf}$
are accumulated only over those~44 instances
where \gringo+\clasp\ never exceeded memory or time limits.

Overall we performed runs for 3 systems,
2 encodings, 4 objectives, and both datasets (54 instances),
which yields 1296 distinct configurations.
We note that memory or time limits (MO and TO)
were always exceeded by the solver,
and never by the grounder.

\myexptable%
The first section of accumulated results
is a comparison of \emph{ASP systems},
i.e., \clingo, \gringo+\clasp, and \gringo+\wasp.
\wasp\ clearly has worse performance in this application
with respect to memory as well as time
compared with \clasp-based configurations.
Columns $T_\mi{grd}$, $\mi{Opt}$, $\mi{Chc}$, $\mi{Cnf}$
\reva{cannot be obtained }for all runs of \wasp\reva{\ due to timeouts},
therefore we omit them in these table rows.
\clingo\ requires more memory than running
\gringo+\clasp\ in a pipe,
moreover, $T_\mi{grd}$ of \clingo\ includes preprocessing time
(we discuss these issues in Section~\ref{secInstaniationIssues}).
In the remaining table we present only results for \gringo+\clasp\ %
\reva{because \wasp\ exceeds the memory and time limit comparatively often.}

In the second accumulation section of Table~\ref{tblExp},
we show a comparison between encodings CM and MM.
Choosing between MM and CM
means a trade-off between time and memory:
while CM exceeds the memory limit less often than MM,
the latter finds optimal solutions for more runs,
and solutions of CM are further away from the optimum
(on average~6.8) in comparison with MM (on average~1.0).

The next section of the table compares objective functions:
voting-based objectives (\objv\ and~\objva)
yield more optimal solutions in a shorter time
compared with the other objectives (\obju\ and~\objua),
and even suboptimal solutions of the former objectives
are often close to optimal
($\mi{Opt}$ is~0.0 on average
for the~44 easiest instances).
Moreover,
strict constraints for transitive links,
as used in~\obju\ and~\objv,
require significantly less memory and instantiation time
compared with weak constraints,
which are used in~\objua\ and~\objva\
for realizing
condition~(C3) of Section~\ref{secObjectives}.

The last section of the table
shows practically relevant \emph{scenarios},
accumulated over single datasets.
For DS1,
non-voting-objectives are practically relevant,
while for DS2 the opposite holds.
DS2 can be automatically adjudicated with the MM encoding
and objective~\objv\ with all optimal solutions in the given time and memory.
For DS1 the most feasible configuration is~\obju\ with encoding CM:
it never exceeds memory but unfortunately also yields no optimal solutions.
In practice, having any solution is better than running out of memory.
Moreover, if we increase resource limits
to~8~GB and~1200~sec (not shown in the table)
then we obtain optimal solutions for all documents of DS1
with~\obju\ and MM,
and suboptimal solutions for all documents with~\objua\ and CM.

To analyze instantiation bottlenecks,
we have modified \gringo~4.5 to print the number of
instantiations of each non-ground rule.%
\footnote{\url{https://github.com/peschue/clingo/tree/grounder-stats}}
For an instance of average difficulty and the UA objective function,
the main instantiation effort of MM encoding is
the transitive closure rule~\eqref{eqMMCCTransitive}
with 725K instantiations,
while for CM it is the weak constraint~\eqref{eqCMTransCost}
for transitivity with 5M instantiations.
These rules clearly dominate over the next most frequently instantiated
rules (12K instances for MM, 112K instances for CM).
Although in encoding CM
we obtain significantly more ground rules than in MM,
the former requires less memory.
A significant difference between the structures of
CM and MM encodings is that
encoding CM is tight \citep{Erdem2003tight}
while encoding MM is not,
due to the transitive closure over predicate $\mi{cc}/2$
in rules~\eqref{eqMMCCCopy}--\eqref{eqMMCCTransitive}.

Note that to the best of our knowledge
there are no other tools for automatic adjudication,
and there are no other published datasets.
Therefore we have no possibility for empirically
comparing our approach with other tools or on other datasets.

\section{Discussion}
\label{secDiscussion}

We have learned the following lessons in this project.

\subsection{Approximation, Modularity, and Declarativity}
The abstract task we solve is quite straightforward.
However, to make its computation feasible,
we need to resort to approximations (as in assuming
a maximum number of chains in the CM encoding),
and we have the possibility to `trade time for space',
just as in classical algorithm development
(choosing between the MM and CM encoding is such a trade-off).

Careful tuning of encodings is necessary.
For example,
replacing constraints~\eqref{eqCMSync1}
and~\eqref{eqCMSync2}
by the following rule and constraint
\begin{align*}
 \mi{clink_{good}}(M_1,M_2)
  &\lars \mi{result_{cm}(C,M_1),\,result_{cm}(C,M_2)},\, M_1 \lts M_2. \\
  &\lars \mi{clink(M_1,M_2),\,\naf\ clink_{good}(M_1,M_2)}.
\end{align*}
makes the encoding perform significantly worse.%
\footnote{Due to symmetry breaking (`$<$') in~\eqref{eqCanonicalize1} and~\eqref{eqCanonicalize2},
this encoding alternative is correct.}
The need for such tuning
makes ASP less declarative
than we would expect (or want) it to be.
Still, the modularity of ASP also facilitates tuning
and finding better formulations:
our encodings share many rules although their
essential way of representing the search space
is very different.

Note that preliminary encodings~\citep{Schuller2016rcraadjudication}
used different objective function formulations,
however, these encodings were not usable in practice
without resorting to aggressive approximations
that degraded results, see also Section~\ref{secInstaniationIssues}.

\subsection{ASP Optimization}

Unsatisfiable-core-based optimization (USC)~\citep{Andres2012}
and stratification~\citep{Alviano2015waspoptimum}
are both essential to the applicability of ASP
in this application.
Obtaining a suboptimal solution is always
better than obtaining no solution at all,
in particular in semi-automatic adjudication.
Branch-and-bound optimization (BB) performed
so much worse than USC in this application
that we omitted any numbers.

We also experimented with
additional symmetry breaking for the CM encoding,
such that solutions with permutated chain IDs
are prevented.
With this extension of the encoding,
we noticed that USC performance was reduced,
while with BB \reva{the frequency of finding better solutions increased}
(although it did not increase enough
to \reva{ reach the performance of USC optimization}).

Experiments with unsat-core-shrinking of WASP
\citep{Alviano2016coreshrinking} have not yielded
better results than with the normal WASP configuration.
Similarly, experiments with lazy instantiation of constraints
\citep{Schuller2016aspfoa,Cuteri2017lazyprop}
have not yielded performance improvements.

\subsection{Instantiation Issues}
\label{secInstaniationIssues}
Analyzing the number of rules instantiated by non-ground
rules can be useful, but it can also be misleading:
in this application the encoding instantiating more constraints (CM) requires less memory in search
(probably because of the tightness of CM).

Another practical issue is,
that measuring instantiation time
with \clingo\ is impossible,
as \clingo\ reports only the sum of
preprocessing and instantiation times.
This makes comparisons with other systems difficult,
hence we opted to compare mainly with \gringo+\clasp.

A small surprising observation is that
\clingo\ consistently requires slightly more memory than \gringo+\clasp.

\section{Related Work}
\label{secRelated}
Our tool is the first automatic tool
for coreference adjudication.
Adjudication is usually performed manually
{\revamark
on a small number of annotations per document and
it is usually performed using graphical interactive tools such as }%
BART~\citep{Versley2008bart},
BRAT~\citep{Stenetorp2012},
or GATE~\citep{Gaizauskas1996,Cunningham2013}.

Our work on automatic adjudication
is {\revamark enabled} by the \reva{number }of annotations
we collected for each document,
which is larger than in usual coreference annotation projects
(we collected at least eight annotations per document)
\reva{and thus }%
permits automatic adjudication under the assumption
that the majority of annotators provides correct annotations.

The computational problem of finding minimal repairs
for inconsistent annotations
is related to finding minimal repairs for
databases~\citep{Chomicki2005} or
ontologies~\citep{Eiter2014dlrepair},
and to managing inconsistency
in multi-context systems~\citep{Eiter2014incmanaij}.
In these problems,
the aim is to find a minimal change in a system
such that the modified system
is globally consistent,
which is similar to our problem of
merging mutually inconsistent coreference annotations
into a single consistent gold standard result.
All these problems have in common,
that a change that fixes one inconsistency
might introduce another one,
potentially in a part of the system that is
not obviously related to the changed part.

Scoring coreference annotations
based on existing and non-existing mention-mention links
(as we do in the \objv\ and \objva\ objectives)
is related to the BLANC~\citep{Recasens2010blanc}
evaluation measure for coreference analysis.
Scoring only based on existing mention-mention links
(as in \obju\ and \objua\ objectives)
is related to the MUC~\citep{Vilain1995muc6coref}
evaluation measure,
which scores precision and recall of
mention-mention links over all gold chains compared
with all predicted chains.
In adjudication, we produce one set of gold chains
from many other gold chains,
while these scores are defined for comparing
pairwise scores between two sets of chains
(one set of predicted chains
and one set of gold standard chains).

ASP encodings for transitivity are present in many applications.
In particular, ASP encodings for acyclicity properties
have been studied by \citet{Gebser2015acyclicity},
who included transitive closure in several
of their encodings.
Similar as in our experiments,
tightness makes a relevant difference.

While coreference resolution is different
from coreference adjudication,
methods related with Answer Set Programming
have been used to perform coreference resolution.
\citet{Denis2009} described an approach for coreference resolution
and named entity classification based on Integer Linear Programming,
which includes transitivity of mention-mention links.
\citet{Inoue2012coref} created an approach for coreference resolution
based on weighted abduction that is evaluated
using an Integer Linear Programming formulation.
Note that Integer Linear Programming is a formalism
that is related to ASP \citep{Liu2012}.
\citet{Mitra2016} describe an approach for question answering
based on Inductive Logic Programming under ASP semantics,
which includes coreference resolution as a subtask.

More remotely related approaches for coreference resolution
include mostly methods for building coreference chains
from scores on mention-mention links
obtained from some machine learning solution on a
labeled coreference corpus.
Such methods include local greedy heuristics
\citep{Bengtson2008,Stoyanov2012},
global optimization formulations
such as relaxation labeling \citep{Sapena2012}
or ranking with Markov Logic \citep{Culotta2007},
and representations of trees of mention-mention links~%
\citep{Fernandes2012system,Chang2013}.
Rule-based algorithms for anaphora resolution
were first described by \citet{Hobbs1978},
more recent systems merge chains in a multi-stage
filtering approach \citep{Lee2013dcoref}.
Hybrid systems combine rules and machine learning
and use curated or distributed knowledge sources
such as WordNet, Google distance, and Wikipedia
\citep{Poesio2004cleanedupentry,Chen2012conll,Zheng2013}.
Most of these systems must be trained on a gold standard corpus
that is the outcome of adjudication
and all of these systems are evaluated on such gold standard corpora.

\section{Conclusion}
\label{secConclusion}
We have developed an ASP application for automatic adjudication of coreference annotations
along with two structurally different real-world benchmark datasets.

As our work is the first automatic adjudication approach,
we could not compare with systems performing the same
or a similar task.
Moreover, there are no standard datasets published
about the problem, because usually only the final result
of adjudication --- the gold standard --- is published.

To solve the automatic adjudication problem
for all practically relevant instances that we encountered,
significant effort and several iterations of encoding improvement were necessary.
We have the impression
that ASP tools do not yet provide sufficient feedback
about problems in encodings
to make the process of encoding optimization an easy task.

Still,
for our datasets, we consider this computational problem as solved,
and we have integrated our encodings
in the publicly available \crasp\ tool
that supports automatic and semiautomatic
adjudication of coreference data in CoNLL-format.

\subsection*{Future Work}

Future work related to ASP %
can be the development of automatic methods for encoding optimization,
similar to initial work by \citet{Buddenhagen2015}.
To be useful, such methods need to be integrated into tools,
in the best case into grounders and solvers that directly give hints how to improve an encoding.

A future topic of research on coreference resolution
and adjudication can be partially or largely overlapping mentions
that are not annotated on exactly the same boundaries.
This issue, in particular, arises with agglutinating
languages such as Turkish
where splitting long words into several tokens
and annotating parts of a token can sometimes be desired.
Currently, in our approach,
as well as in common coreference scoring schemes,
overlapping and non-overlapping mentions are treated equally as if they are completely distinct mentions.

\section*{Acknowledgements}
We are grateful for the feedback of
reviewers at \reva{the RCRA workshop,
at the LPNMR conference,
and at the Journal of Experimental \& Theoretical Artificial Intelligence}.

\section*{Funding}
This work has been supported by
The Scientific and Technological Research Council of Turkey
(TUBITAK)
under grant agreements 114E430 and 114E777.

\printbibliography

\end{document}